\documentclass[conference]{IEEEtran}
\IEEEoverridecommandlockouts

\usepackage{LA}
\usepackage{algos}

\usepackage{cite}
\usepackage{amsmath,amssymb,amsfonts}

\usepackage{graphicx}
\usepackage{textcomp}
\usepackage{xcolor}
\def\BibTeX{{\rm B\kern-.05em{\sc i\kern-.025em b}\kern-.08em
    T\kern-.1667em\lower.7ex\hbox{E}\kern-.125emX}}

\usepackage{cleveref} 
\usepackage{booktabs}

\begin{document}

\title{Inverting Neural Networks: New Methods to Generate
        Neural Network Inputs from Prescribed Outputs\\
\thanks{This work was supported in part by the National Science Foundation under Grant No. 1949230.}}

\author{
    \IEEEauthorblockN{
           Rebecca Pattichis\IEEEauthorrefmark{1},
           Sebastian Janampa\IEEEauthorrefmark{2},
           Constantinos S. Pattichis\IEEEauthorrefmark{3}, and 
           Marios S. Pattichis\IEEEauthorrefmark{2}}
    \IEEEauthorblockA{
    	\IEEEauthorrefmark{1}Department of Electrical Engineering,
	University of California, Los Angeles, Los Angeles CA, USA \\
    	Email: rebeccapattichis2000@gmail.com}
    \IEEEauthorblockA{
    	\IEEEauthorrefmark{2}Department of Electrical and Computer Engineering,
    	University of New Mexico, Albuquerque, NM, USA\\
    	Emails: \{sebasjr1966, pattichi\}@unm.edu}
    \IEEEauthorblockA{
    	\IEEEauthorrefmark{3}Department of Computer Science,
    	University of Cyprus, Cyprus\\
    	Email: pattichis.costas@ucy.ac.cy}
    }

\maketitle

\begin{abstract}
Neural network systems describe complex mappings that
           can be very difficult to understand.
In this paper, we study the inverse problem of determining the
           input images that get mapped to specific neural network
           classes.
Ultimately, we expect that these images contain recognizable
           features that are associated with their corresponding
           class classifications.
We introduce two general methods for solving the inverse problem.
In our forward pass method, we develop an inverse method based
           on a root-finding algorithm and the Jacobian with respect to
           the input image.
In our backward pass method, we iteratively invert each layer, at the top.
During the inversion process,
          we add random vectors sampled from the null-space of
          each linear layer.
We demonstrate our new methods on both transformer architectures
          and sequential networks based on linear layers.
Unlike previous methods, we show that our new methods are able
          to produce random-like input images that yield near perfect
          classification scores in all cases, revealing vulnerabilities in the underlying networks.          
Hence, we conclude that the proposed methods provide a more comprehensive
          coverage of the input image spaces that solve the inverse mapping problem.            
\end{abstract}


\section{Introduction}
Neural network (NN) systems are used
     to represent complex mappings from inputs to outputs.
For image classification problems, a NN system
     maps high-dimensional image data to a low-dimensional 
     space of class probabilities.
Understanding and interpreting such complex mappings requires that
     we also study the inverse problem: what types of images are
     mapped to specific classes (e.g., see \cite{pattichis2024understanding}).

Solutions to the inverse problem can provide great insights about
      the stability of the NN system.
For example, by examining the input images for a specific class, we
      hope to see that the system relies on reasonable image features.
If not, then it would be possible to attack the network by using 
      unrelated inputs that yield specific classes (e.g., see 
      \cite{goodfellow2014explaining}).
Furthermore, for classification problems,
      we may want to use undesirable input images that yield specific
      classes to retrain the network to classify them correctly 
      (e.g., adversarial learning \cite{lowd2005adversarial}).            
On the other hand, when we return
      desirable images, 
      we can use the inverse map as an image generator.
 
The paper studies the inverse problem of determining
    NN inputs that yield a specific output distribution, building on prior research \cite{pattichis2024understanding}.
In \cite{pattichis2024understanding}, we studied the use of
    vector spaces for understanding NN representations
    and inverting their outputs.
Earlier work on inverting NNs using trainable input images
    was described in \cite{saliency2014}.
Notably, our methods do not require additional training 
    (e.g., decoders).

This paper makes several contributions over prior methods.
First, we present a forward pass algorithm
     that uses random sampling and accelerated root-finding methods
     based on the Jacobian of the NN with respect to the input.
We note that the forward pass algorithm can be applied to any
    NN architecture block.     
Second, we present a more sophisticated backward pass algorithm
     that can be applied to NN architecture blocks that
     consist of \texttt{softmax}, invertible activation functions (e.g., \texttt{SELU}),
     and linear layers.
Our backward pass algorithm inverts each layer sequentially and
     utilizes random sampling from the null spaces of the linear layers
     to generate sample images.
Third, we demonstrate our methods on networks based on
     linear layers as well as well as modern transformer networks.
Our results expose network vulnerabilities by producing 
    random-like images that yield near perfect classification 
    scores.


\section{Background}\label{sec:background}
Earlier research on the use of vector spaces for understanding
       linear layers is based on \cite{pattichis2024understanding}.
In \cite{pattichis2024understanding}, we defined the null-space of a linear layer
       as the rejected signal space.
Here, we will use random samples from the rejected signal space to generate sample
       outputs for the prior layer.
       
In \cite{pattichis2024understanding}, we also considered
    two approaches for calculating input images for prescribed outputs.
Our first method searched the training dataset 
    to find the an input image that maximized our desired class probability.
For our first method, we also computed the average of all images
     that fell within the lowest 25th percentile of distances to the desired 
     (ideal) output values.
This first method, however, is not able to search the entire 
    space of possible images.
Our second approach trained the image so as to minimize
     the mean squared error (e.g., see \cite{saliency2014}).
For this second method, we use the first method to initialize the search.

A fundamental limitation of these prior approaches is that the generated
     images are close to the original training set.
In the current paper, we will show that our proposed methods will generate
     random-like input images that are far from the training set.   

\section{Methods}\label{sec:methods}
For our methods, we assume that the NN has been pre-trained for a specific problem.
We describe two different methods below. 

\begin{figure}[!t]
\caption{\label{algo:nnforward} Neural Network inversion using forward pass.} 
\begin{algorithmic}[1] 
\onehalfspacing
\Function{ForwardPassInv}{NN, ClassDistr, Loss}
\commtop The function generates inputs for any given ClassDistr.
\commtop The method has no restrictions on the Neural Network.
\commtop  It can be used with \textsc{BackPassInv(.)}.
\inp
\commvar NN represents the neural network model.
\commvar ClassDistr describes the desired output prob. distr.
\commvar Loss function describes the loss function to minimize.
\outp
\commvar x represents a random sample image so that:
\commvar \qquad NN(x) outputs ClassDistr
\spc
\State \textbf{Sample} a random initial guess for $x_0$.
\State \textbf{Compute} Jacobian $J$ for loss function wrt x.
\State $x \gets \text{RootFinding}(x_0, J)$
\spc
\State \textbf{Verify} NN(x) generates ClassDistr
\State \textbf{return} x or error
\EndFunction
\end{algorithmic}
\singlespacing
\end{figure}

\subsection{Forward Pass Algorithm}
Our forward pass algorithm (Fig. \ref{algo:nnforward}) relies on the use
       of traditional root-finding methods for generating inputs
       (e.g., see \cite{nocedal2006numerical}).
We start with a random input vector $x$ that
       is used as an initial guess in a root-finding algorithm.
Root-finding algorithms solve the general
       problem of computing $y = F(x)$ for multidimensional
       inputs and outputs.
To accelerate the convergence, we 
       supply the algorithms with the Jacobian
       of the loss function used to train the NN.
Provided that the root-finding algorithm starts sufficiently
       close to a solution, we can guarantee convergence
       \cite{nocedal2006numerical}.
In practice, by randomizing the initial guess image,
      we expect the root-finding algorithm to
      converge to different local solutions.
       
\subsection{Backward Pass Algorithm}

We summarize the algorithm in Fig. \ref{algo:nnbackward}. For the example in Fig. \ref{algo:nnbackward},
      we consider a classification network given by:
\begin{equation}
   y = \texttt{softmax} \left ( 
   	   \phi_n \left(
	      \dots \,
	       \phi_{2} \left ( 
	             W_2 \phi_1 \left (  W_1 x + b_1 \right ) + b_2
	             \right )
	       \right )    
           \right )
           \label{eq:genform}
\end{equation}
where: 
\begin{equation*}
\begin{aligned}
   \phi_1, \dots, \phi_n &\quad \text{are invertible activation functions}, \\
   W_1, \dots, W_n     &\quad \text{are the weight matrices, and} \\
   b_1, \dots, b_n       &\quad \text{are the bias terms for the linear layers}.  
\end{aligned} 
\end{equation*}
The algorithm works by inverting
  backwards from the \texttt{softmax} layer to layer $n$, down to layer $1$.

We invert the \texttt{softmax}  layer (up to a constant) using:
\begin{equation*}
    \begin{aligned}
        \texttt{softmax}^{-1}(y_i) = \log(y_i) - \frac{1}{c} \sum_{j=1}^c \log(y_j) 
    \end{aligned}
\end{equation*}
where $y$ is our desired class probability distributions, or  
    \texttt{ClassDistr} in Figures \ref{algo:nnforward} and 
    \ref{algo:nnbackward}, and $c$ is the number of output classes in our 
    classification model.
       
We define an activation function $\phi_i$ to be invertible
   provided that $\phi_i$ is one-to-one so
    that $\phi_i^{-1} ( \phi_i (y)) = y$ for any
    $y \in D_i$. 
Examples of invertible activation functions include the 
    \texttt{LeakyRelu} and \texttt{SELU}.    
Unfortunately, numerically, inverting activation functions can
   be unstable for specific input-output values.
For example,  both activation
    functions can have relatively large condition numbers
    for vectors that contain positive and negative numbers for
    \texttt{LeakyRelu}  and for large negative numbers for \texttt{SELU}.
As a result, to avoid instability in the calculations, in cases
    where the inversion leads to an unbounded number (large number),
    we return an error message.          
   
For inverting $x_i^- = W_i x_{i-1} + b_i$ to get $x_{i-1}$, 
    we use the Singular Value Decomposition (SVD)
    \cite{trefethen2022numerical}.
To avoid instability, we modify the relative condition
    number of the pseudoinverse $A^+$ as follows.
Let the SVD decomposition for $W_i$ be 
    $W_i = U_i \Sigma_i V_i^T$.
The pseudoinverse is  then given by    
   $W_i^+ = V_i \Sigma_i^+ U_i^T$ where
   $\Sigma_i^+$ contains $1/\sigma_i$ along
   the diagonal.
The relative condition number $\kappa$ is given by:
   $\kappa = \sigma_1/\sigma_k$,
   where $\sigma_1$ denotes the largest singular value and
   $\sigma_k$ denotes the smallest non-zero singular value.
Large values for $\kappa$ signify instability in the inversion process.
To avoid instability, for large $\kappa$, we zero out 
   the lowest singular values by a threshold representing a reasonable condition number maximum, such as 100
   (line \ref{alg:cond} in the algorithm).
As a result, the new $\kappa$ can be significantly reduced at
   the expense of computing a stable approximation to the inverse
   (see line \ref{alg:inv1} in the algorithm).    

A key innovation of our approach is that we use the null-space
   of $W_i$ to generate all possible inputs $x_{i-1}$ that can be used
   to generate $x_i^-$.
This is accomplished by creating a vector of random coefficients,       multiplying it by the $\text{Null}(W_i)$ basis vectors, and then
  normalizing the output using the norm of the SVD inverse vector
  (lines
  \ref{alg:sample} and \ref{alg:null} of the algorithm).    
The process is then repeated for each layer.
At the end,  the learned image is substituted back into the neural
   network to verify that it produces the given probability distribution.
Otherwise, it returns an error.

A key property of \textsc{BackPassInv}(.) is that it generates 
    all possible input images that can yield the given class distribution.
On the other hand, \textsc{ForwardPassInv}(.) relies on random input sampling
    to discover new inputs.    

\begin{figure}[!t]
\caption{\label{algo:nnbackward} Neural Network inversion using backward pass.} 
\begin{algorithmic}[1] 
\onehalfspacing
\Function{BackPassInv}{NN, ClassDistr, Std, CondNum}
\commtop The function generates inputs for any given ClassDistr.
\commtop The method is restricted to classification problems 
\commtop using SoftMax, invertible activation functions, 
\commtop and linear layers. It can be used with
\commtop  \textsc{ForwardPassInv(.)}.
\inp
\commvar NN represents the neural network model.
\commvar ClassDistr describes the desired output prob. distr.
\commvar Std controls the variance for drawing samples from 
\commvar \qquad the Null-space.
\commvar CondNum is an optional threshold for 
\commvar \qquad controlling the condition of the weight matrices.
\outp
\commvar x represents a random sample image so that:
\commvar \qquad NN(x) outputs ClassDistr
\spc
\State \textbf{Invert} \texttt{SoftMax} layer by 
\State \qquad computing inputs $x_{n}$ from ClassDistr.
\For{layer $i \gets n, n-1, \dots, 1$}
\State \textbf{Invert} the activation function $\phi_i$ to get $x_i^-$.
\State \textbf{If} $x_i^-$ grows unbounded \textbf{return} error.
\State \textbf{Stabilize} $W_i^+$ to CondNum maximum.            \label{alg:cond} 
\State $x_{i-1} \gets W_i^+(x_i^- - b_i)$.                                      \label{alg:inv1}
\State \textbf{Sample} random null-space vector $S$ using Std. \label{alg:sample}
\State $x_{i-1} \gets x_{i-1} + \text{Null}(W_i) S. $                    \label{alg:null}
\EndFor
\spc
\State \textbf{Verify} NN(x) generates ClassDistr
\State \textbf{return} x or error
\EndFunction
\end{algorithmic}
\singlespacing
\end{figure}
    
\subsection{Generalized Method} 
Consider a general Neural Network (NN) form given by:
\begin{equation*}
   \text{NN} (x) = \text{NN}_n (\dots\, \text{NN}_2 (\text{NN}_1 (x)) )
\end{equation*}
where $\text{NN}_1, \text{NN}_2, \dots, \text{NN}_n$ denote 
    neural network architecture blocks.
Here, we note that each block operates on the output of the previous
    block as given by $\text{NN}_i (\text{NN}_{i-1} (.))$.     
By using the mean-squared error loss function, we can clearly use 
    the \textsc{ForwardPassInv}(.) to invert each block.
Furthermore, any block that consists of invertible activation functions 
   and linear layers, with or without \texttt{softmax}, can be inverted
   using the methods described in \textsc{BackPassInv}(.). 
Thus, while \textsc{ForwardPassInv}(.) generates a single input,
   a combination of \textsc{ForwardPassInv}(.) and \textsc{BackPassInv}(.)
   can be used to generate a distribution of random input samples.   

\section{Results}\label{sec:results}

\subsection{\textsc{ForwardPassInv} Applied to Transformer Architectures}
We consider the application of the forward pass to transformer architectures used in image
       classification.
Here, we note that Vision Transformers (ViT) only pass the class embedding 
       to the classification layers.
We run our experiments on two different ViT models, tiny-ViT \cite{vit} and 
    DINOv3-Base \cite{dinov3}, fine-tuned on the MNIST dataset, achieving validation accuracy of 95.7\% and 99.8\%, respectively.
The models specifications and training hyper-parameters are summarized in 
    Table \ref{tab:training_params}. 
We use the \texttt{timm} library \cite{timm} and its FlexiViT \cite{flexivit} 
    implementation to resize the patch embedding kernel to $4\times4$. 
We use the Levenberg-Marquardt algorithm as the root-finding algorithm 
       \cite{nocedal2006numerical}.
We apply up to 30 iterations and verify convergence of the algorithm.
       
\begin{table}[h!]
	\centering
	\caption{Model parameters, training hyper-parameters}
	\begin{tabular}{l | c c}
		\toprule
		& ViT-tiny & DINOv3-Base\\
		\midrule
		patch size & $4\times4$ & $4\times4$\\
		input channels & 1 & 1\\
		hidden dim & 192 & 768 \\
		num. of layers & 12 & 12 \\
		epochs & 32 & 32 \\
		batch size & 32 & 32 \\
		EMA model & True & True\\
		EMA warm up iters & 500 & 500\\
		optimizer & AdamW & AdamW \\
		\bottomrule
	\end{tabular}
	\label{tab:training_params}
\end{table}

\subsection{Backward Pass Algorithm Applications}
To test the backward pass algorithm, we train fully-connected neural 
    networks (FCNN) with varying sizes, namely 1, 2, and 6 linear layers. 
    Each Linear layer was followed by an SELU activation function, although 
    any invertible activation function can be used (e.g., Leaky 
    ReLU).
Each FCNN is trained with the SGD optimizer, learning rate of 
    0.001, momentum of 0.9, and 20 epochs, matching the 
    hyperparameters from \cite{pattichis2024understanding}.
The 1-, 2-, and 6-layer FCNNs get 91.5\%, 95.4\%, and 96.9\% validation accuracy, 
    respectively.
During the inversion process, we set \texttt{Std} to 0.1.

\begin{figure*}
	\centering
	\resizebox{\textwidth}{!}{
		\begin{tabular}{c cccccccccc}
            \rotatebox{90}{train-data}&
			\includegraphics[width=0.09\textwidth]{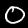} & \includegraphics[width=0.09\textwidth]{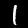} &
			\includegraphics[width=0.09\textwidth]{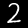} & \includegraphics[width=0.09\textwidth]{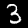} &
			\includegraphics[width=0.09\textwidth]{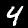} & \includegraphics[width=0.09\textwidth]{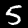} &
			\includegraphics[width=0.09\textwidth]{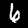} & \includegraphics[width=0.09\textwidth]{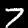} &
			\includegraphics[width=0.09\textwidth]{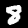} & \includegraphics[width=0.09\textwidth]{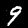}
			\\

            \rotatebox{90}{\cite{pattichis2024understanding}-1layer}&
			\includegraphics[width=0.09\textwidth]{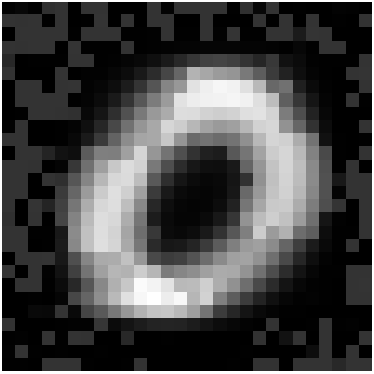} & \includegraphics[width=0.09\textwidth]{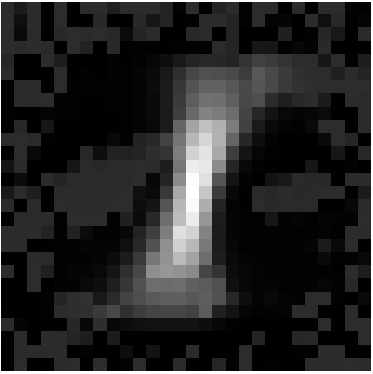} &
			\includegraphics[width=0.09\textwidth]{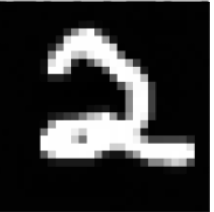} & \includegraphics[width=0.09\textwidth]{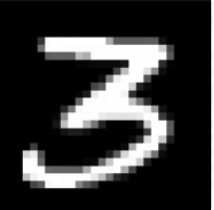} &
			\includegraphics[width=0.09\textwidth]{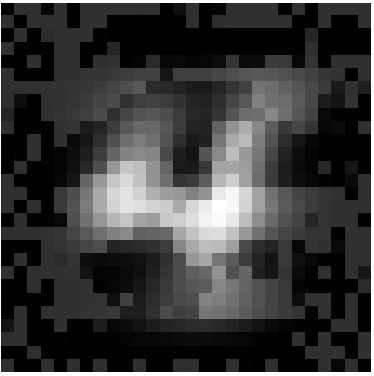} & \includegraphics[width=0.09\textwidth]{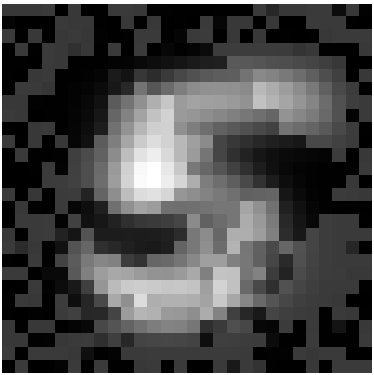} &
			\includegraphics[width=0.09\textwidth]{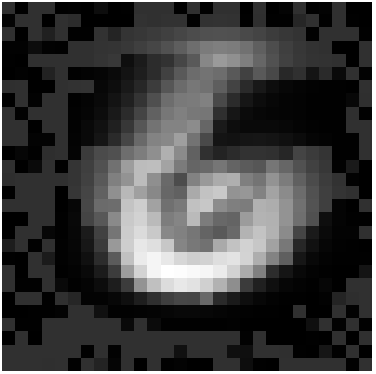} & \includegraphics[width=0.09\textwidth]{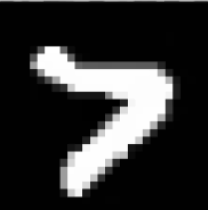} &
			\includegraphics[width=0.09\textwidth]{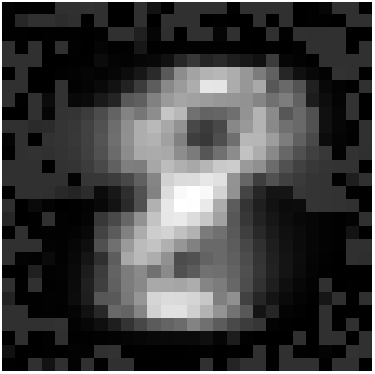} & \includegraphics[width=0.09\textwidth]{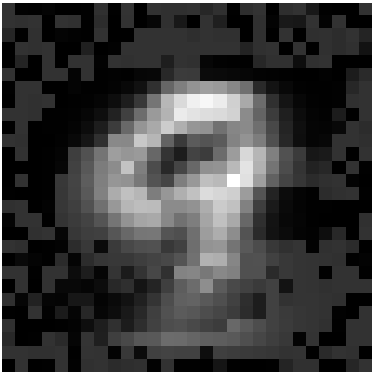}
			\\

            \rotatebox{90}{1layer-NN}&
			\includegraphics[width=0.09\textwidth]{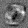} & \includegraphics[width=0.09\textwidth]{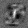} &
			\includegraphics[width=0.09\textwidth]{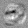} & \includegraphics[width=0.09\textwidth]{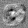} &
			\includegraphics[width=0.09\textwidth]{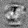} & \includegraphics[width=0.09\textwidth]{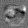} &
			\includegraphics[width=0.09\textwidth]{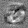} & \includegraphics[width=0.09\textwidth]{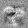} &
			\includegraphics[width=0.09\textwidth]{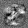} & \includegraphics[width=0.09\textwidth]{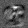}
			\\

            \rotatebox{90}{2layer-NN}&
			\includegraphics[width=0.09\textwidth]{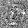} & \includegraphics[width=0.09\textwidth]{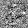} &
			\includegraphics[width=0.09\textwidth]{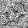} & \includegraphics[width=0.09\textwidth]{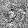} &
			\includegraphics[width=0.09\textwidth]{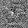} & \includegraphics[width=0.09\textwidth]{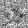} &
			\includegraphics[width=0.09\textwidth]{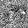} & \includegraphics[width=0.09\textwidth]{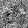} &
			\includegraphics[width=0.09\textwidth]{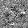} & \includegraphics[width=0.09\textwidth]{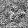}
			\\

            \rotatebox{90}{6layer-NN}&
			\includegraphics[width=0.09\textwidth]{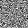} & \includegraphics[width=0.09\textwidth]{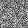} &
			\includegraphics[width=0.09\textwidth]{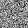} & \includegraphics[width=0.09\textwidth]{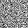} &
			\includegraphics[width=0.09\textwidth]{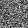} & \includegraphics[width=0.09\textwidth]{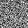} &
			\includegraphics[width=0.09\textwidth]{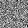} & \includegraphics[width=0.09\textwidth]{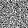} &
			\includegraphics[width=0.09\textwidth]{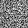} & \includegraphics[width=0.09\textwidth]{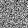}
			\\
            
			\rotatebox{90}{ViT-tiny}&
			\includegraphics[width=0.09\textwidth]{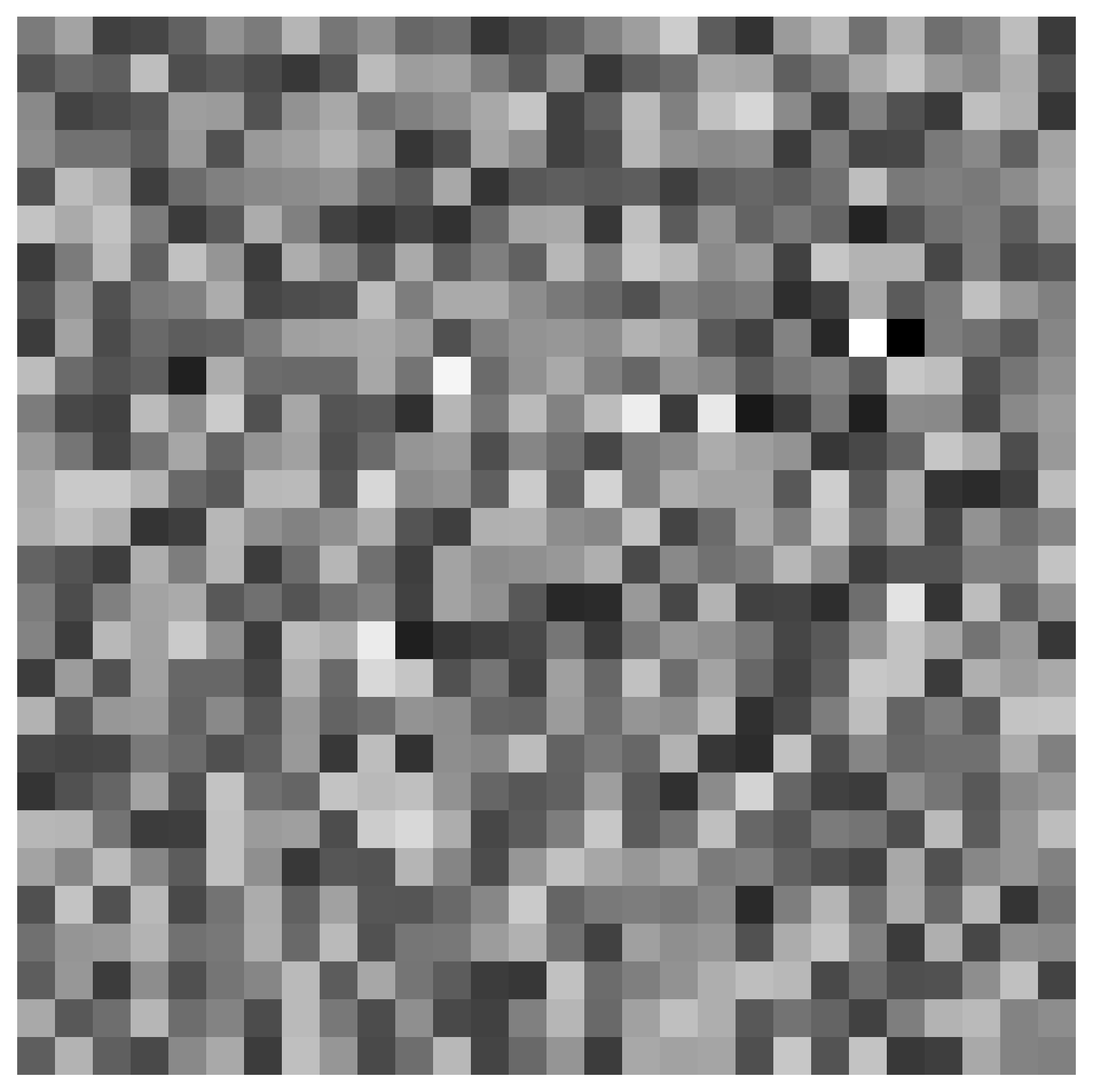} & \includegraphics[width=0.09\textwidth]{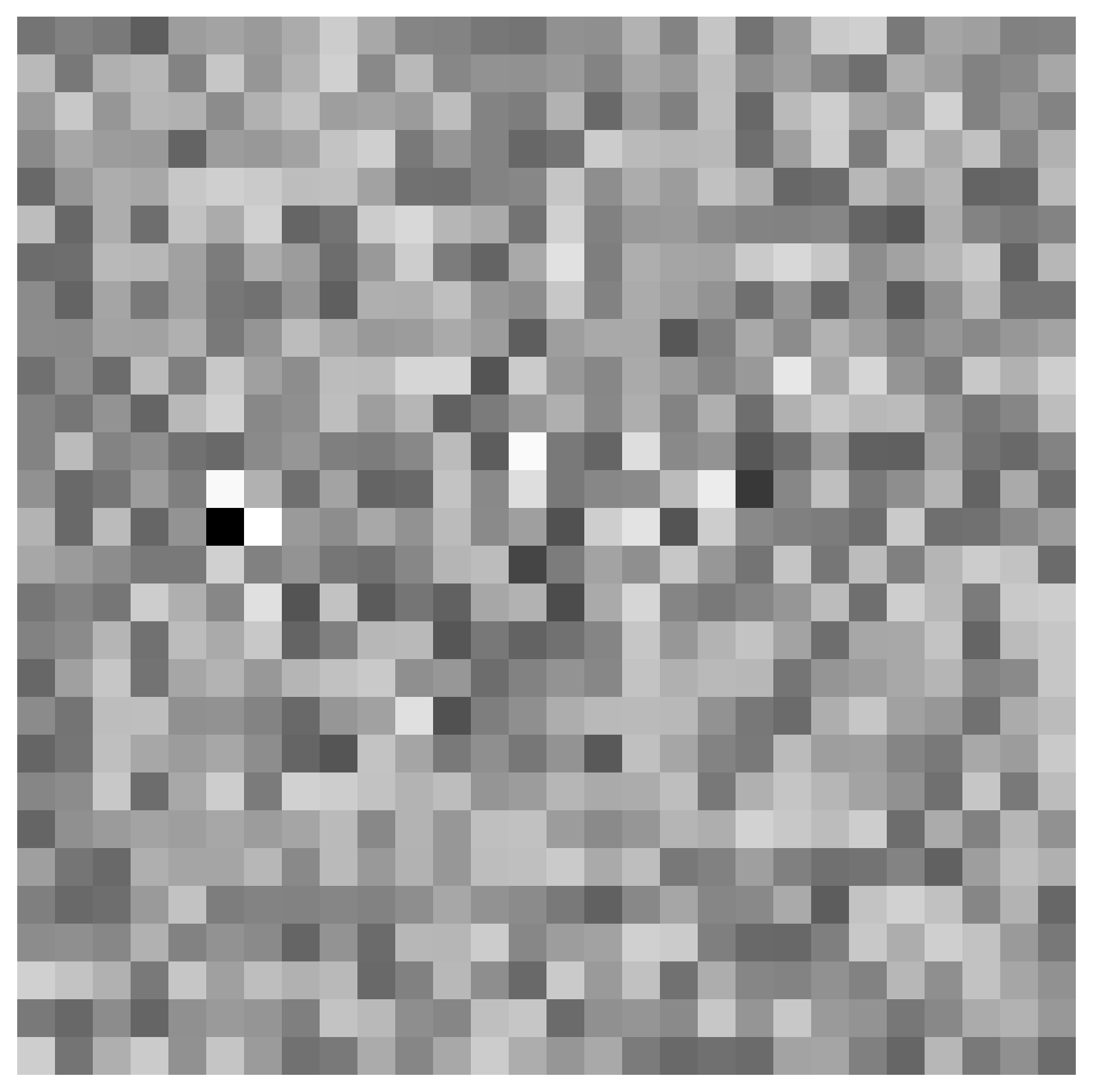} &
			\includegraphics[width=0.09\textwidth]{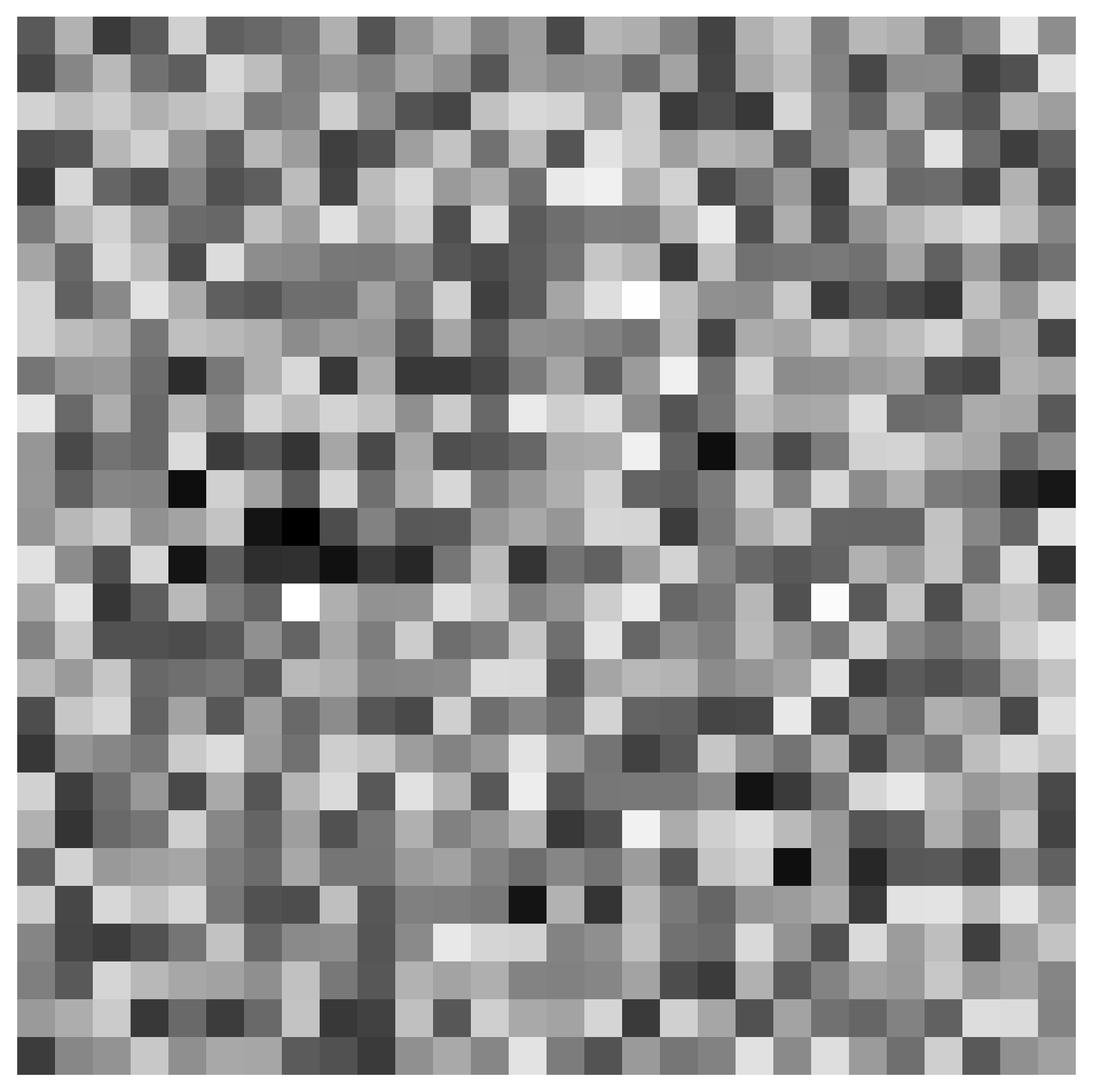} & \includegraphics[width=0.09\textwidth]{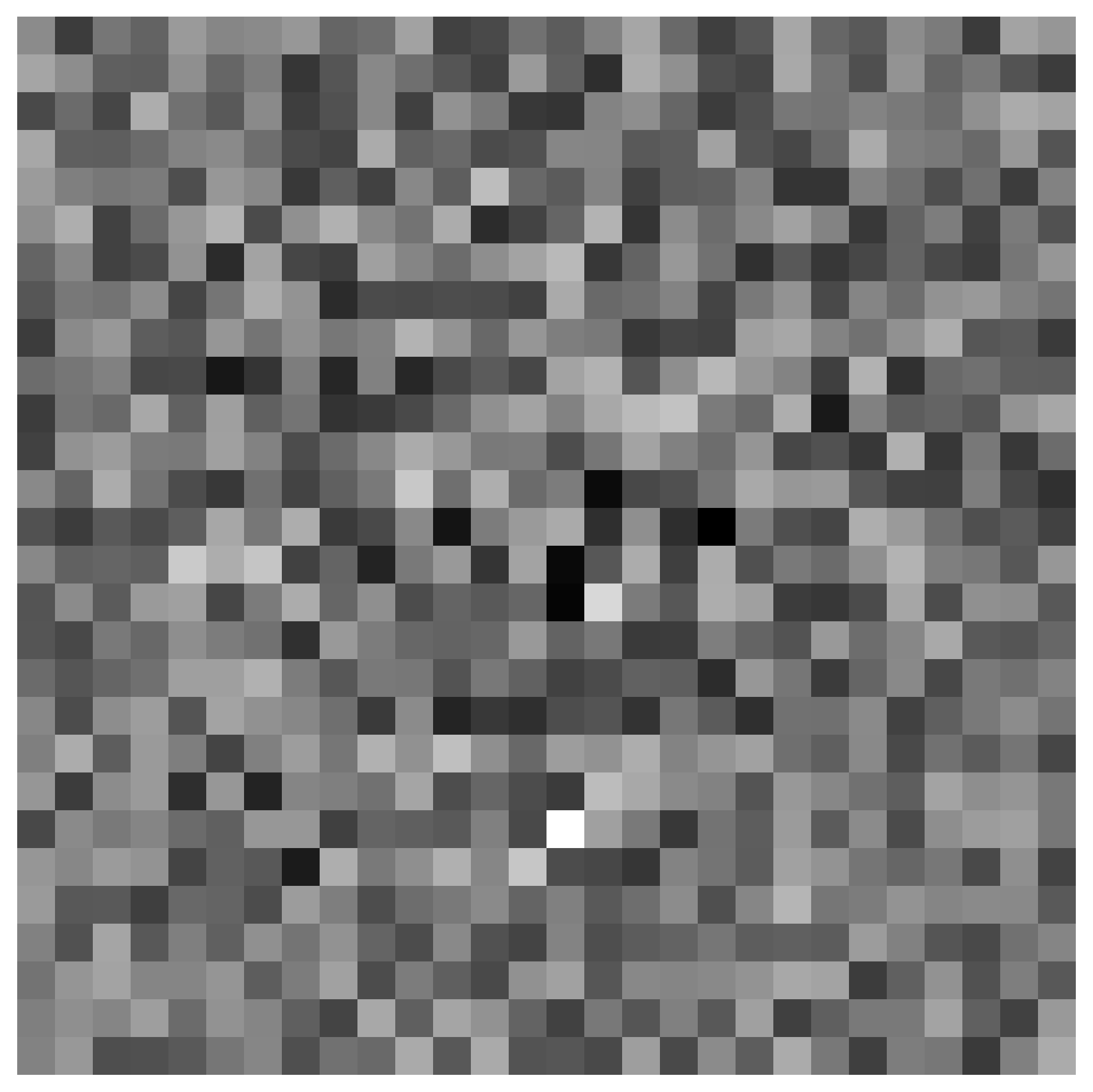} &
			\includegraphics[width=0.09\textwidth]{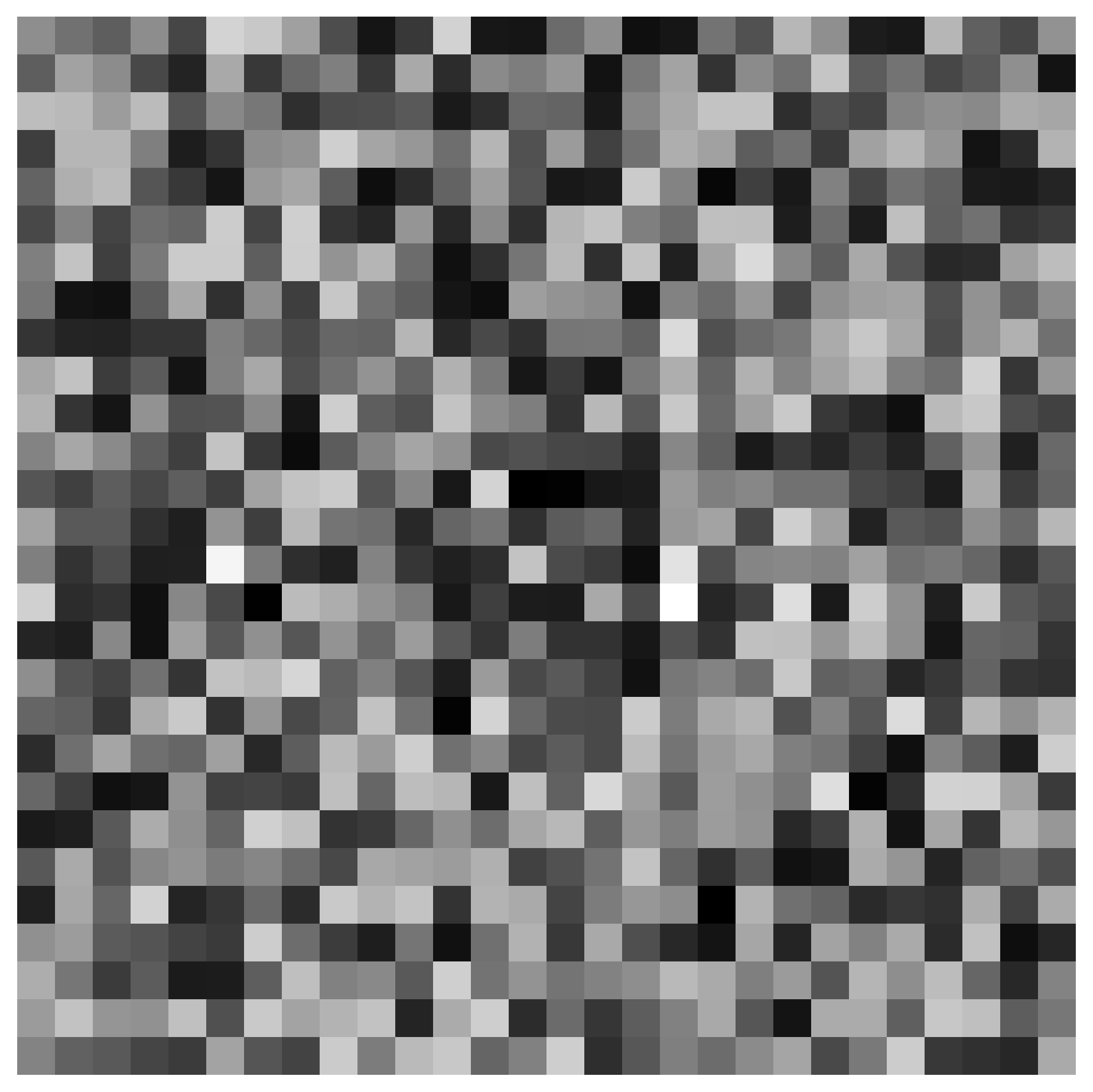} & \includegraphics[width=0.09\textwidth]{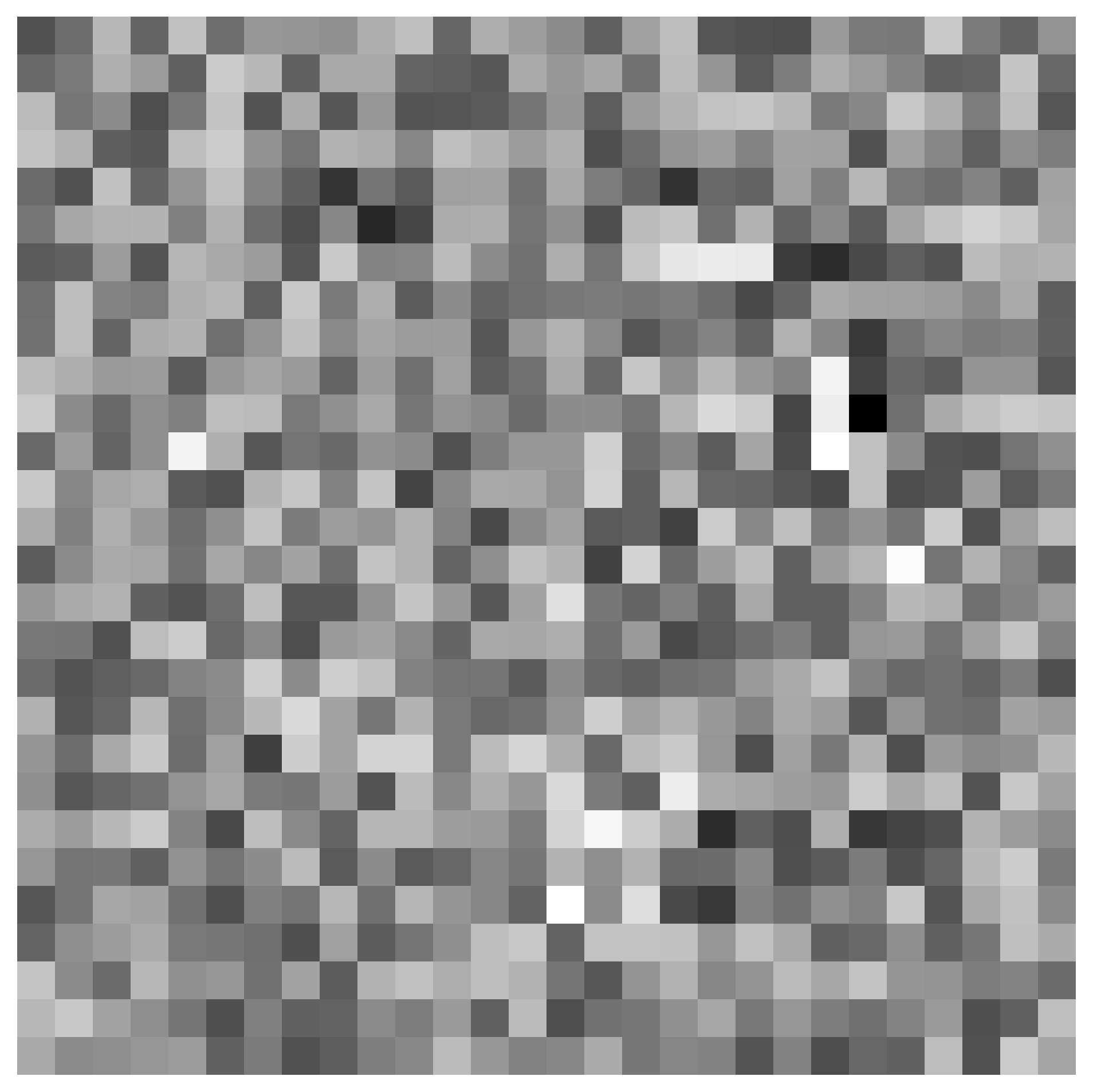} &
			\includegraphics[width=0.09\textwidth]{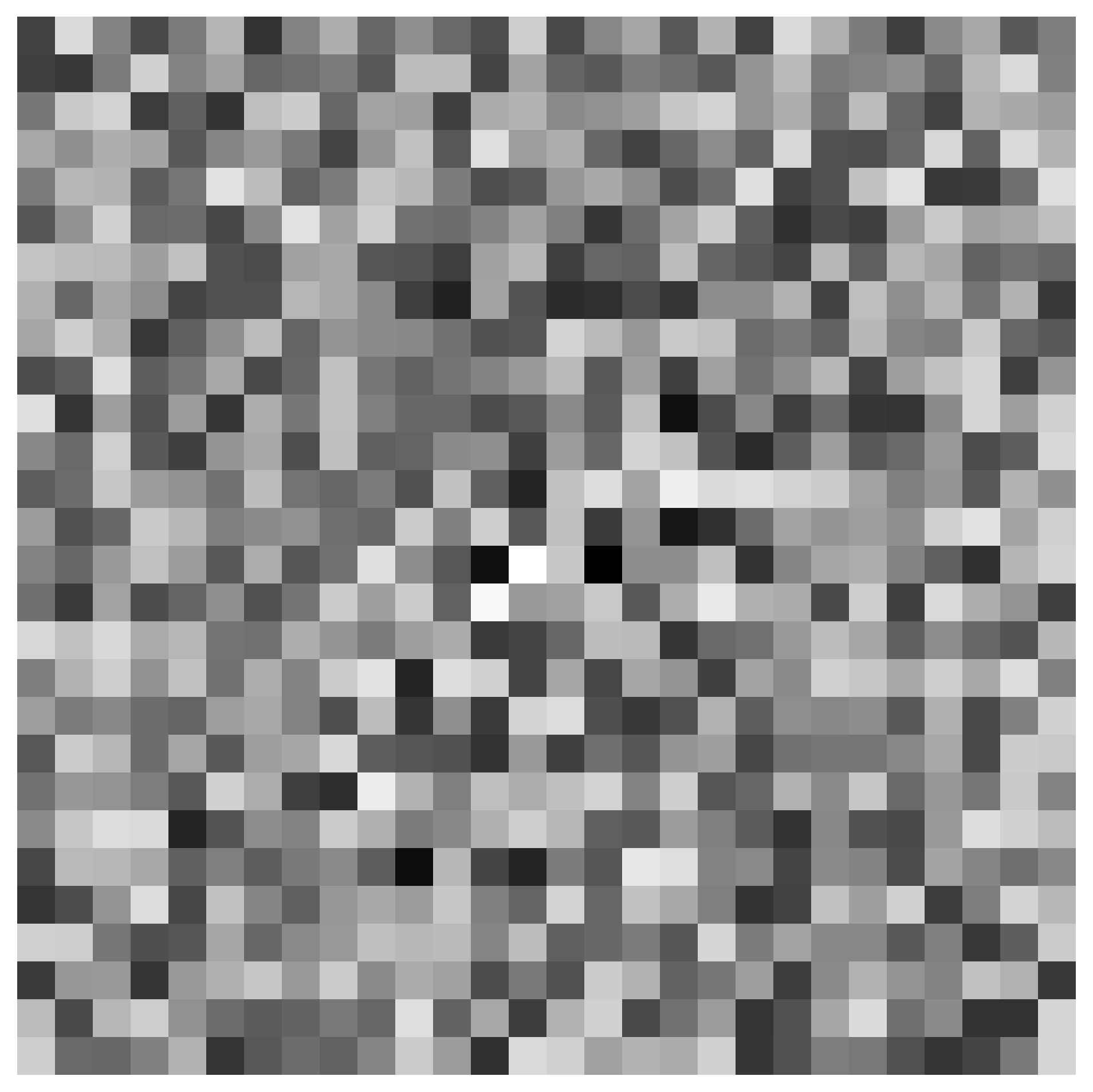} & \includegraphics[width=0.09\textwidth]{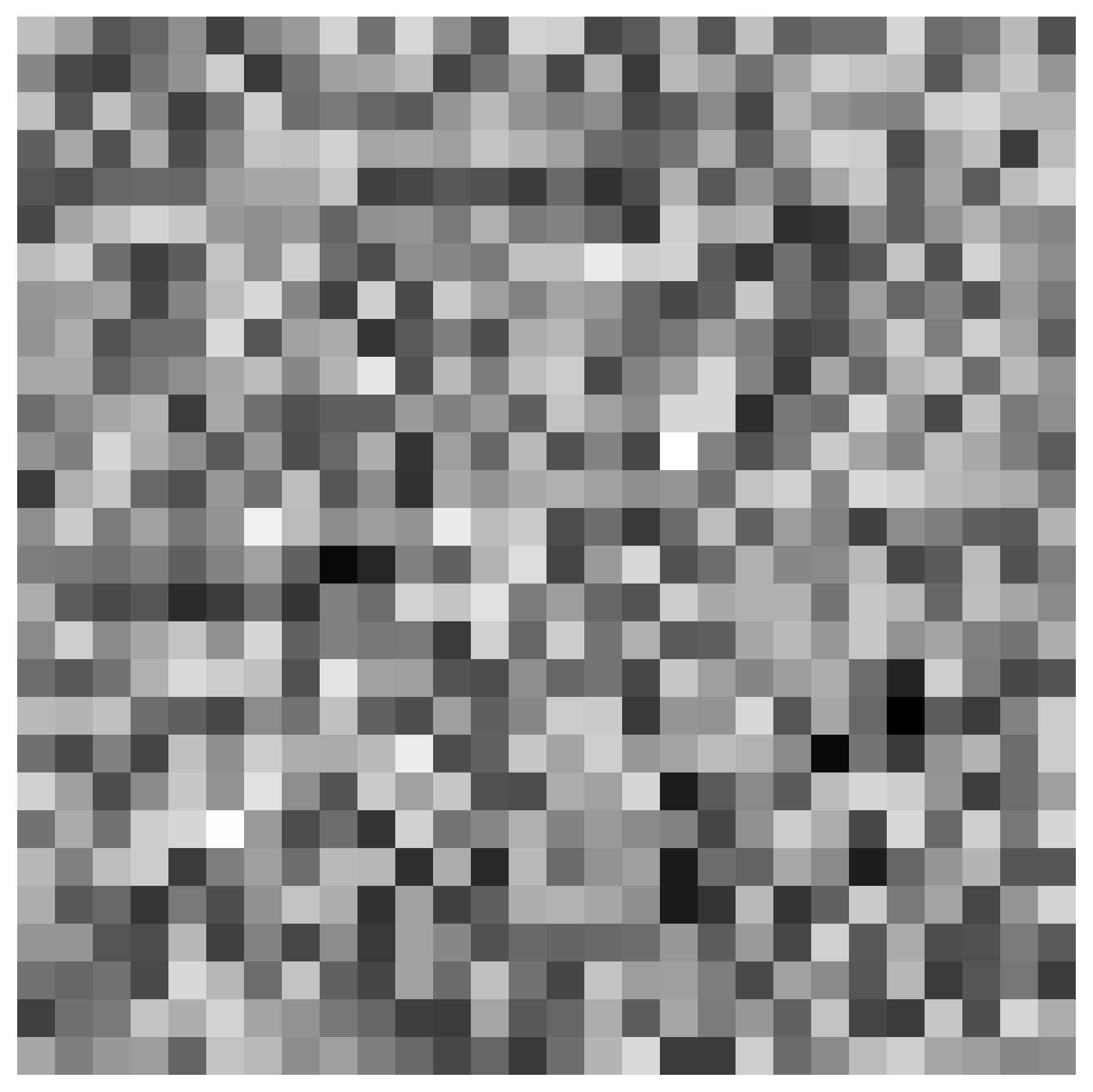} &
			\includegraphics[width=0.09\textwidth]{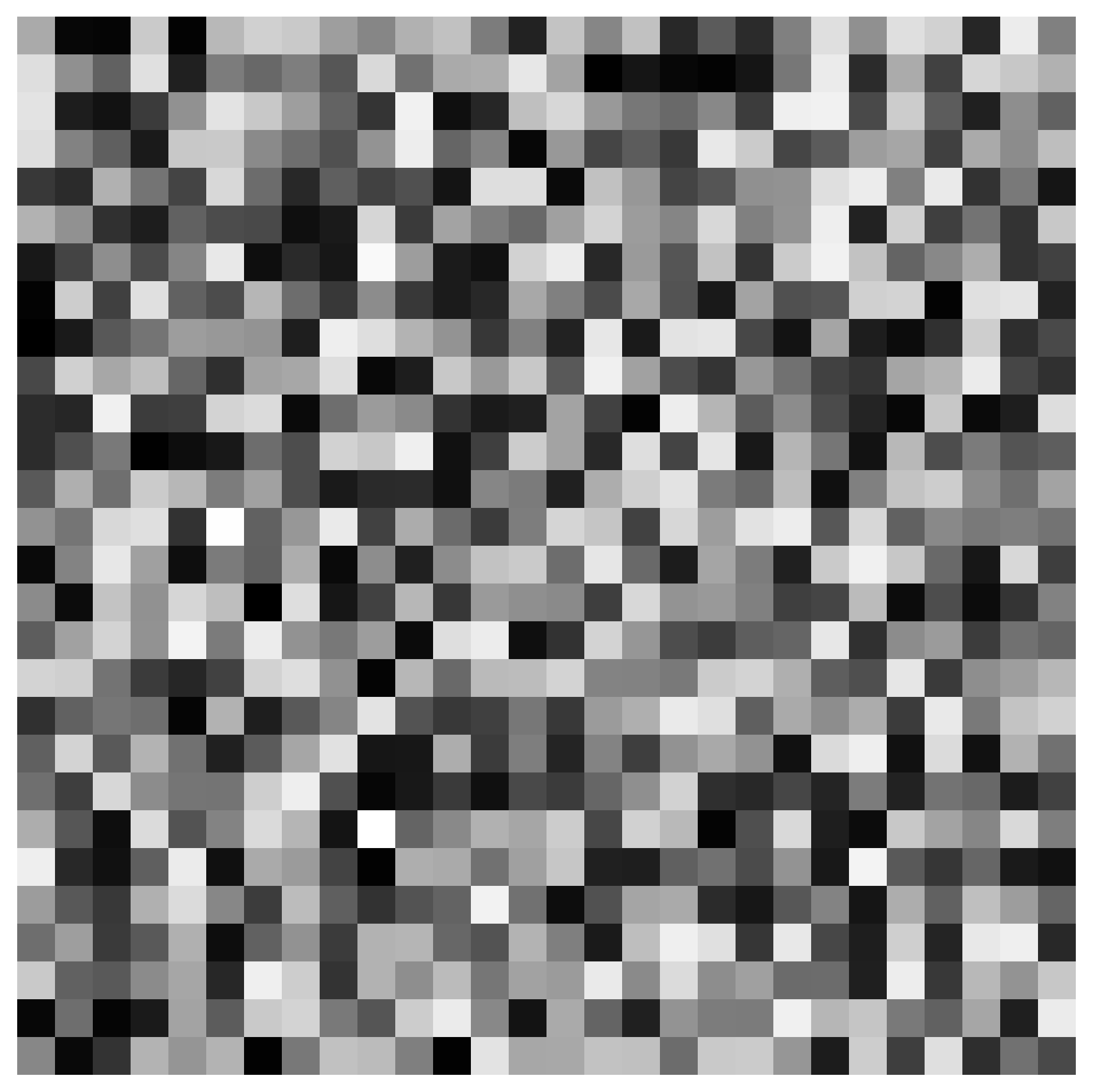} & \includegraphics[width=0.09\textwidth]{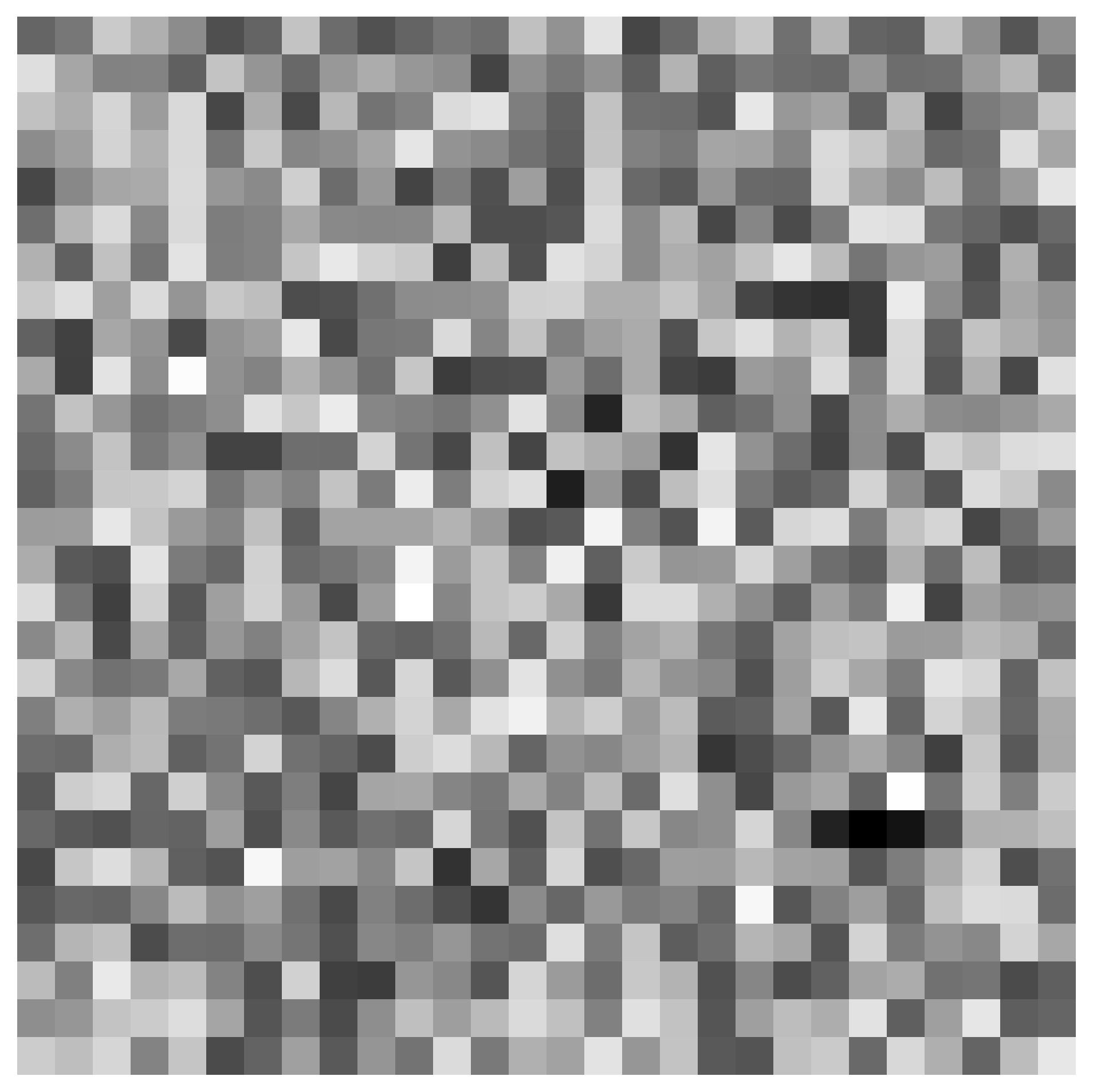}
			\\
			
			\rotatebox{90}{DINOV3} &
			 \includegraphics[width=0.09\textwidth]{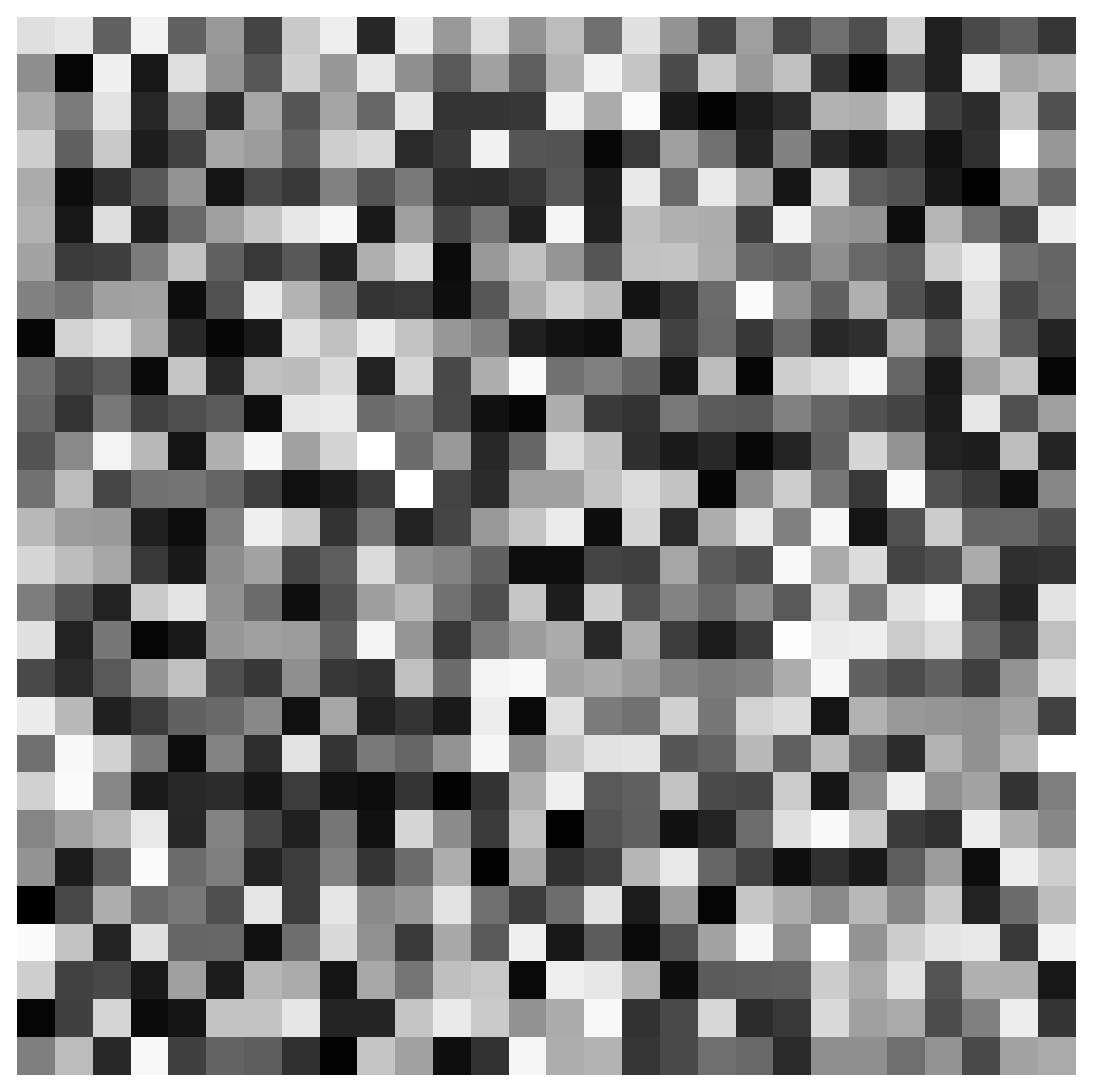} & \includegraphics[width=0.09\textwidth]{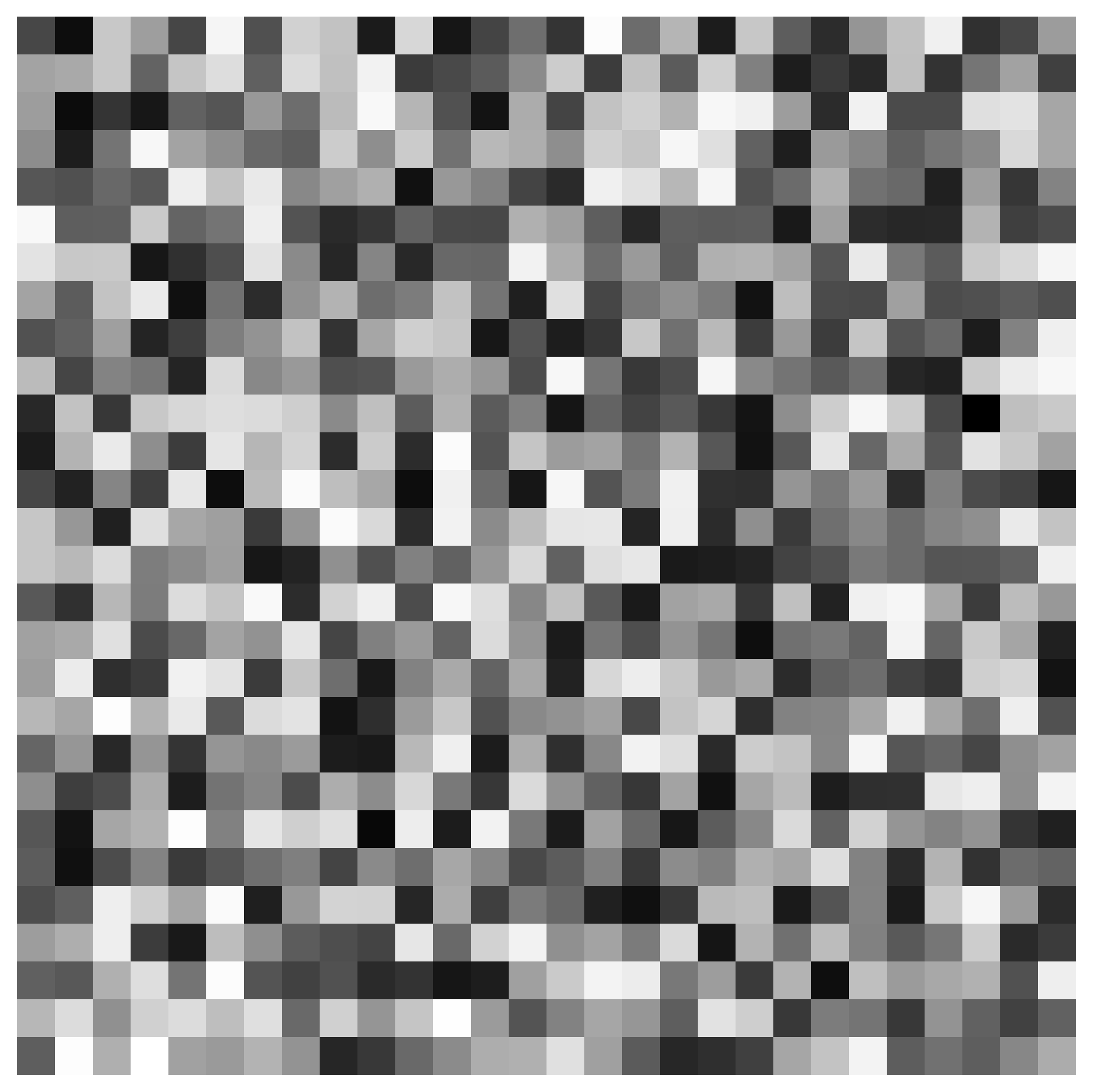} & 
			 \includegraphics[width=0.09\textwidth]{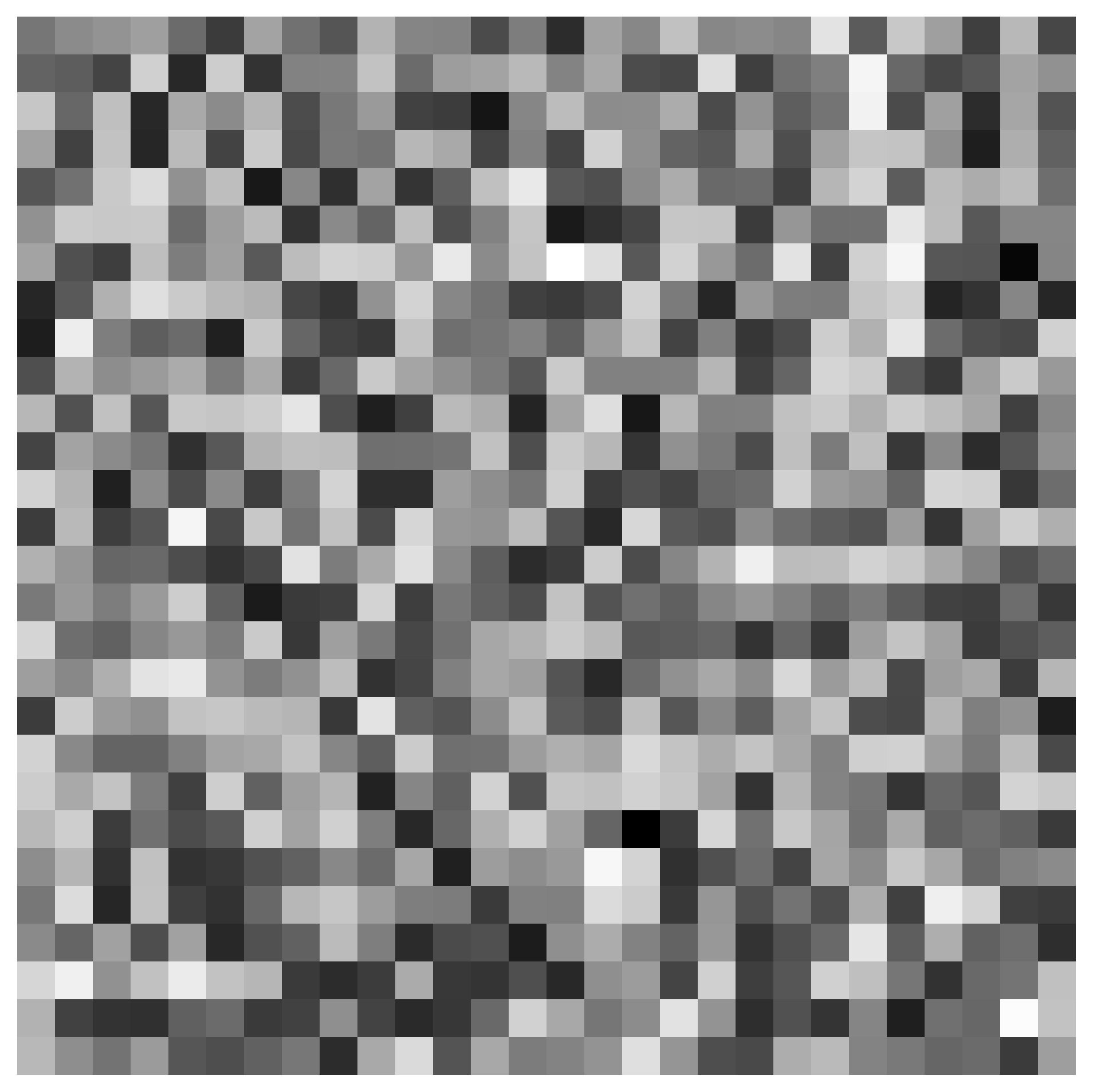} & \includegraphics[width=0.09\textwidth]{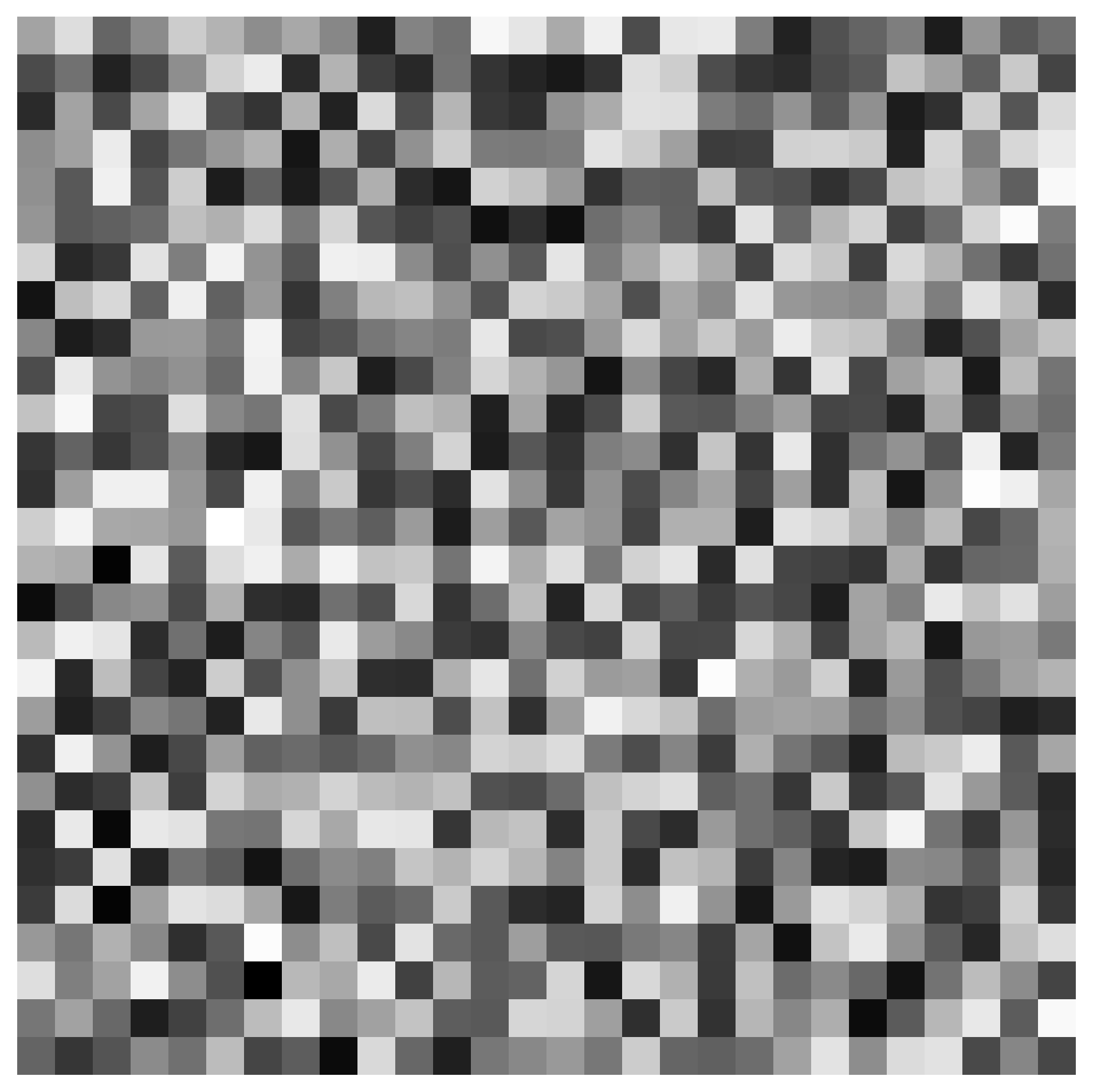} & 
			 \includegraphics[width=0.09\textwidth]{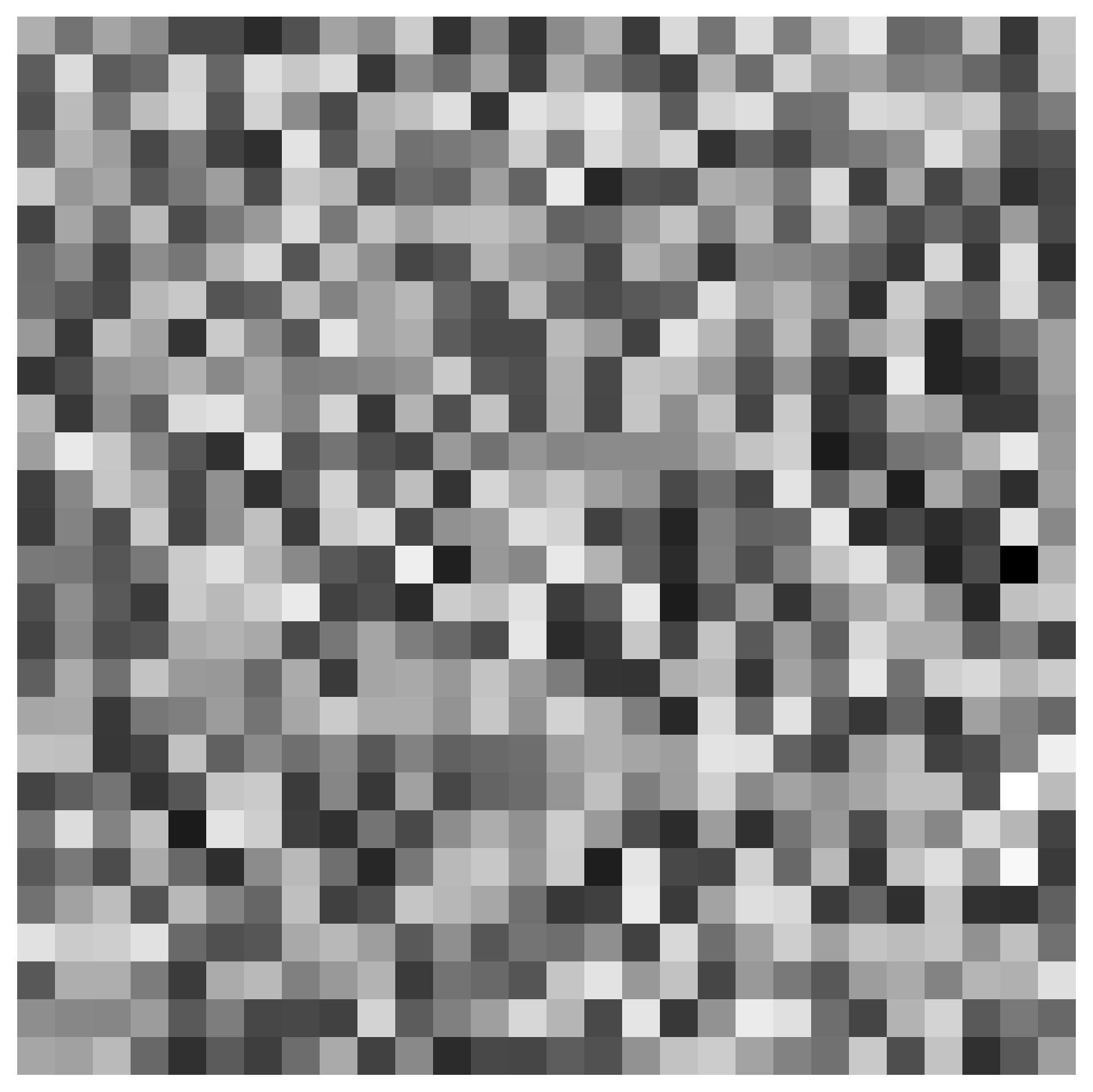} & \includegraphics[width=0.09\textwidth]{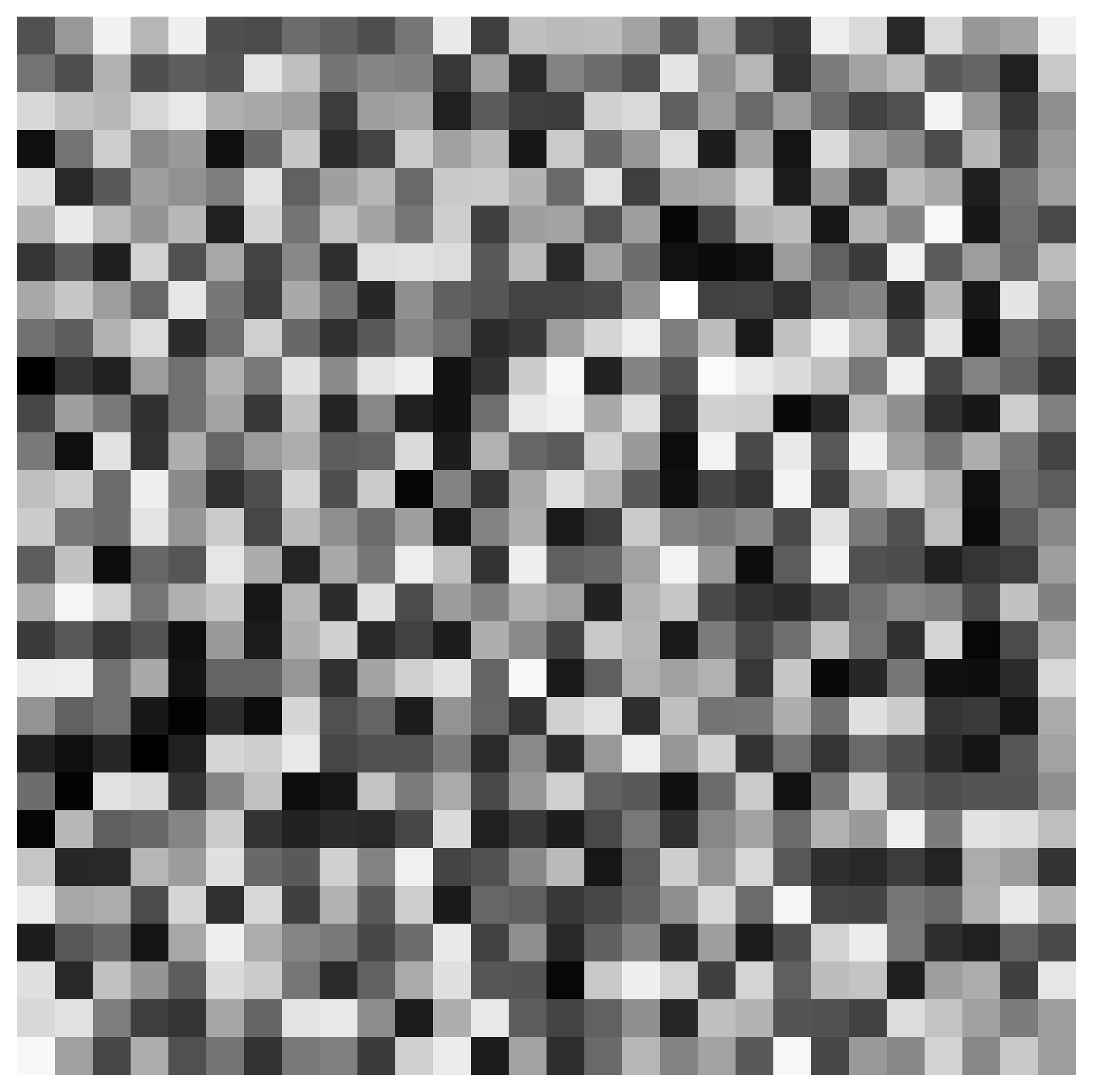} & 
			 \includegraphics[width=0.09\textwidth]{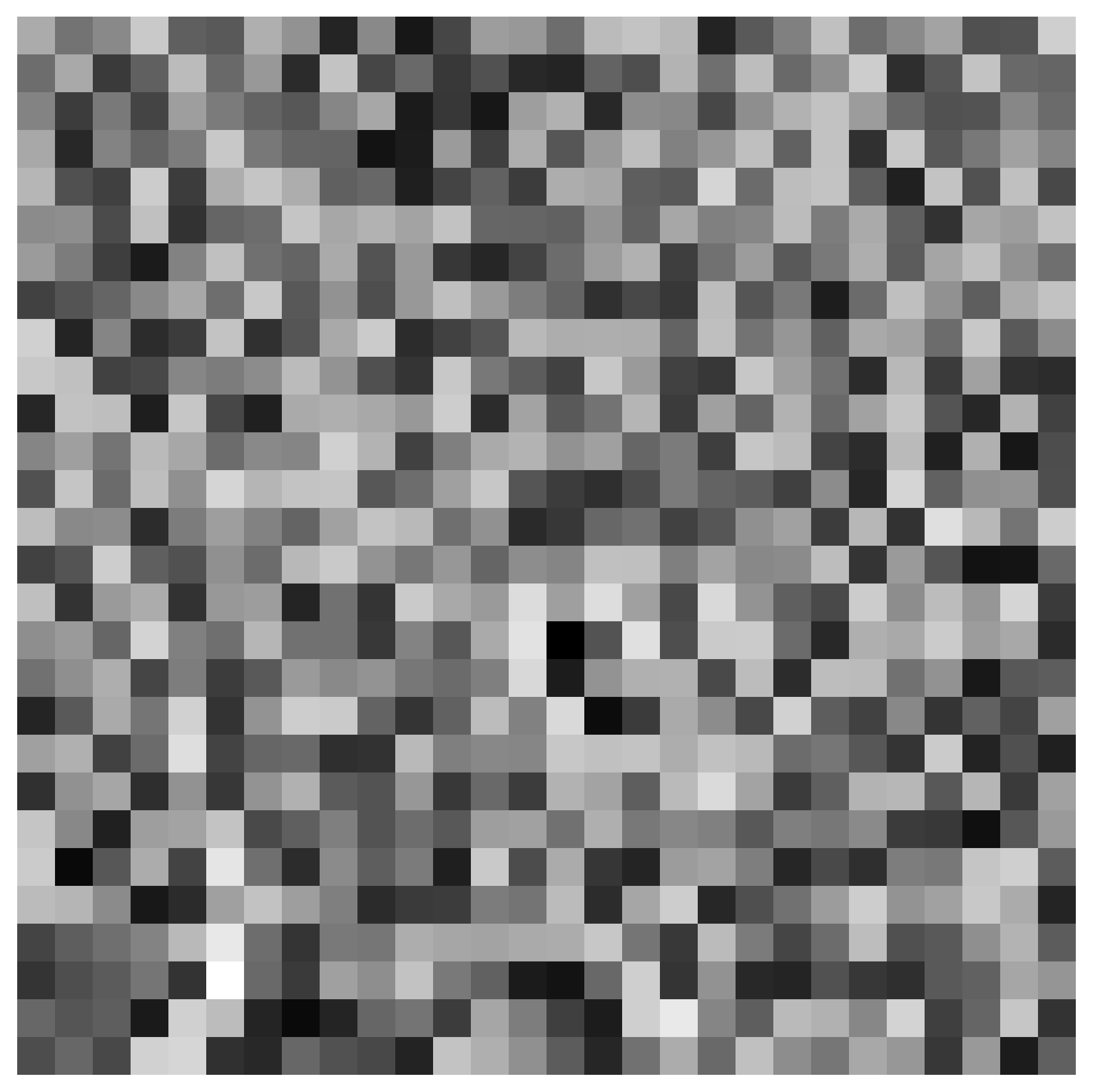} & \includegraphics[width=0.09\textwidth]{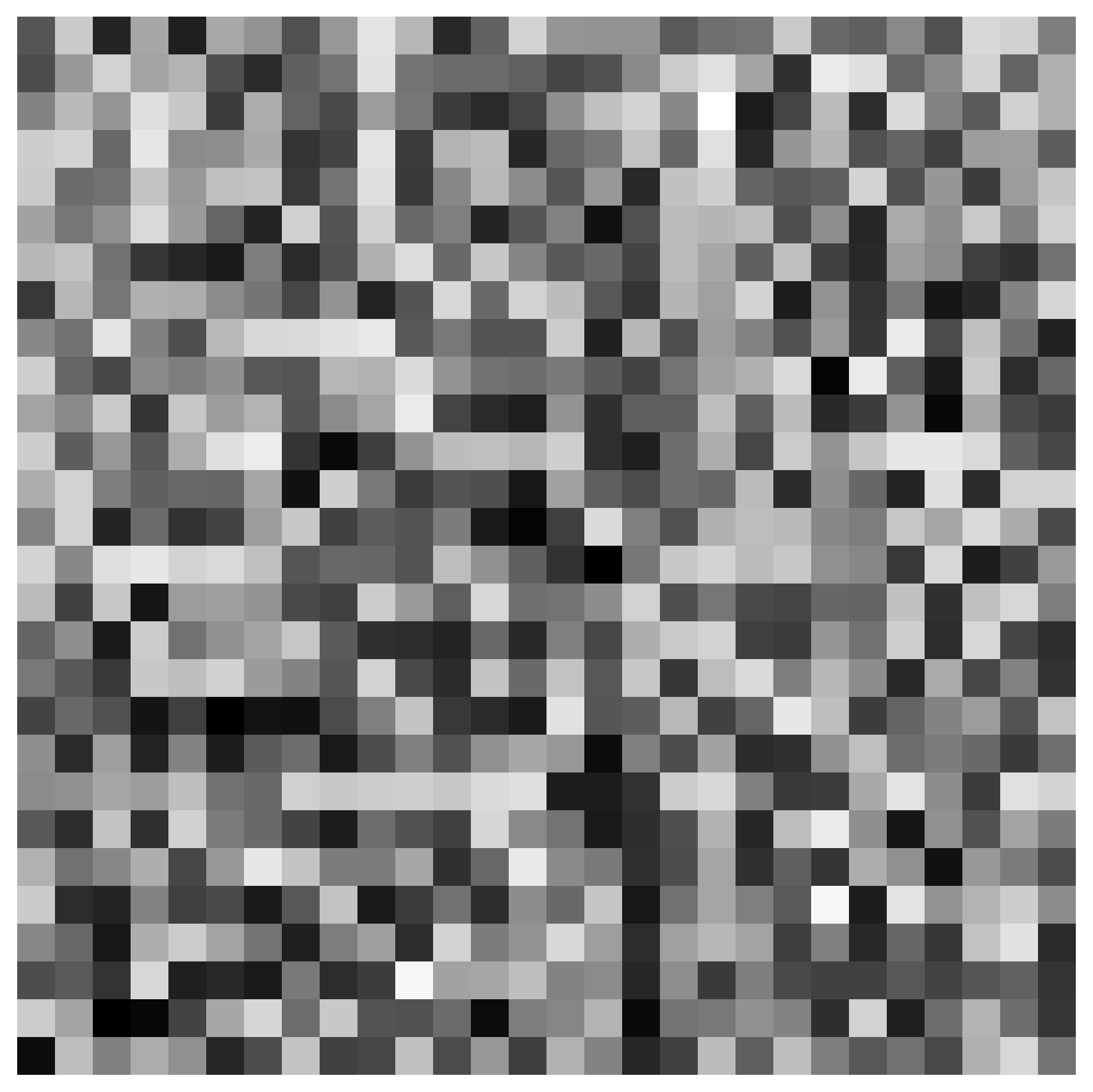} & 
			 \includegraphics[width=0.09\textwidth]{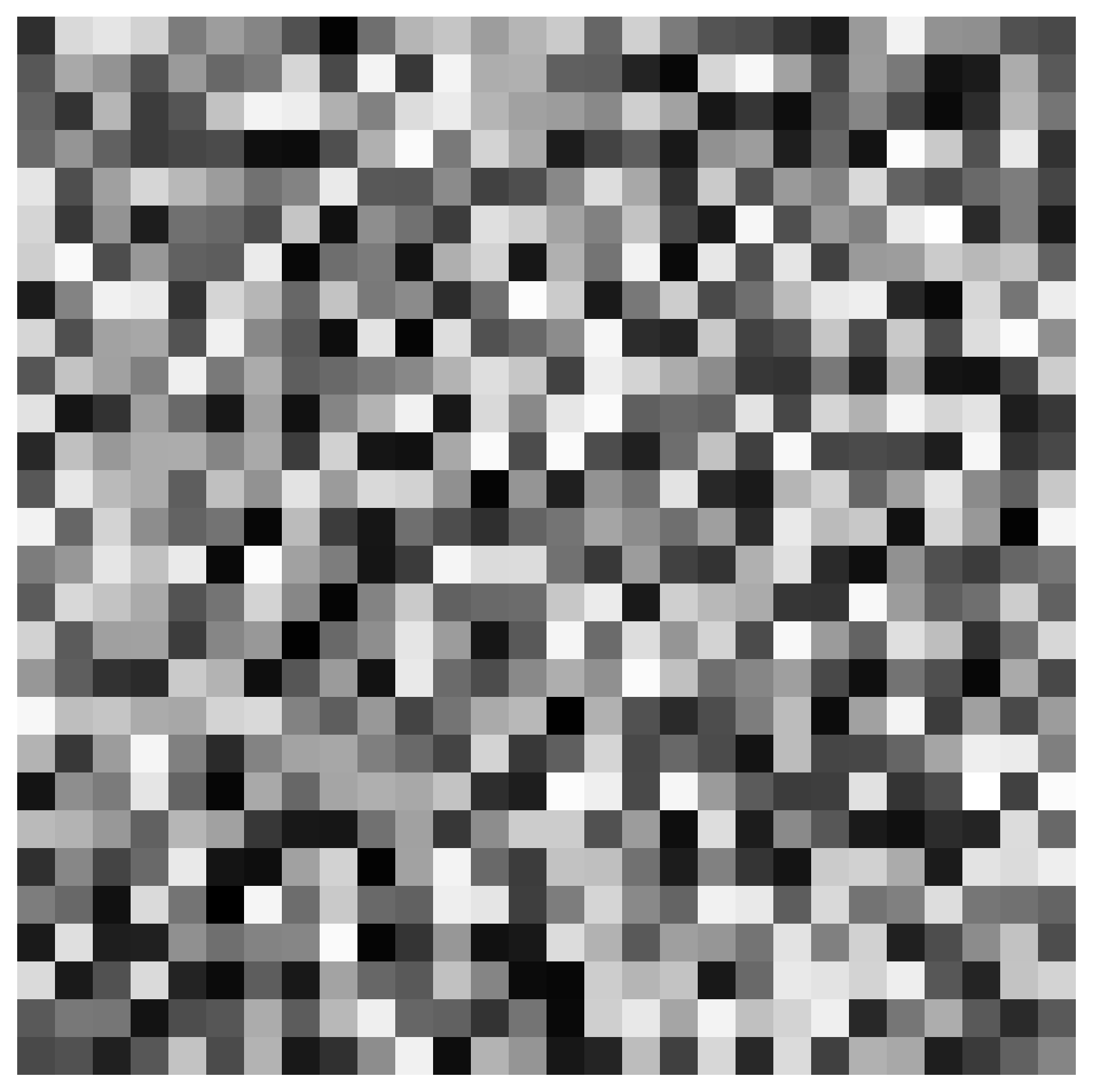} & \includegraphics[width=0.09\textwidth]{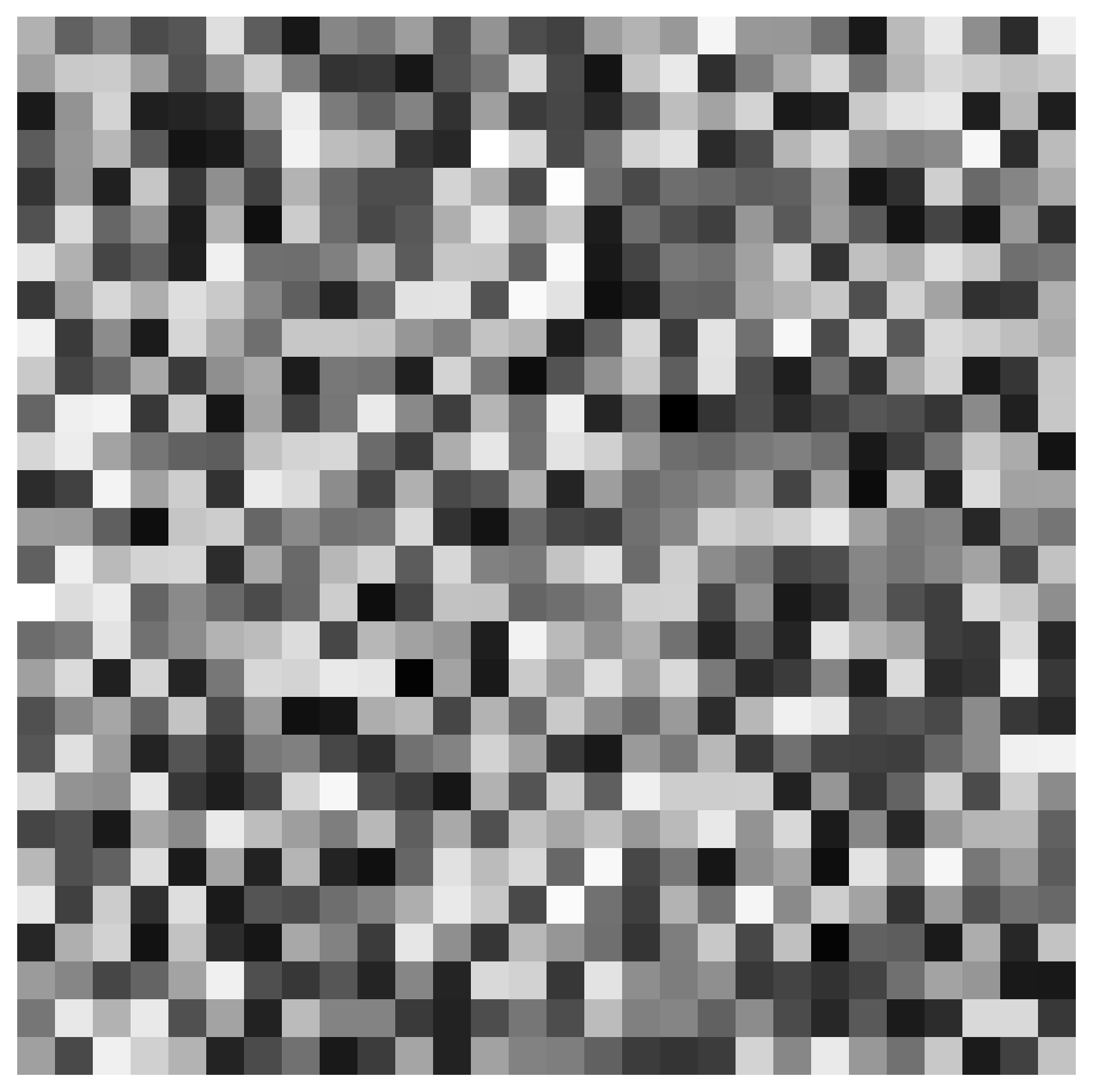} \\
			& Class 0 & Class 1 & Class 2 & Class 3 & Class 4 & Class 5 & Class 6 & Class 7 & Class 8 & Class 9 
		\end{tabular}
	}
	\caption{Generated ideal input images using different algorithms. 
    \textbf{train-data row:} reference images given in the MNIST training data for each class. 
    \textbf{\cite{pattichis2024understanding}-1layer row:} ideal images generated in \cite{pattichis2024understanding}. \textbf{1layerNN, 2layerNN, and 6layer NN rows:} ideal images generated \textsc{BackPassInv} algorithm on FCNNs, with null-space noise added ($\texttt{Std}=0.1$). \textbf{ViT-tiny and DINOV3 rows:} generated images using \textsc{ForwardPassInv}(.). All generated images from new methods produce a probability score $\geq0.9$ for each digit class and $\leq$0.1 for the other ones.}
	\label{fig:learned_images}
\end{figure*}

\subsection{Discussion}
We provide comparative results in Fig. \ref{fig:learned_images}.
The first row displays sample images from the MNIST training set.
From our prior work in \cite{pattichis2024understanding}, second row in Fig. \ref{fig:learned_images}, we can see
       that the previous algorithms gave images that were very close to the training dataset.
Following, our new methods generate images that give $\geq$ 0.9 probability
       for the target class and $\leq 0.1$ probability for the remaining digits.

Our new methods provide random-like input images that expose the 
       vulnerabilities of the underlying networks.
It is clear from our results that image randomness increases dramatically
       with the complexity of the network.
For example, while the 1-layer images of the third row bear some resemblance
       to the training set images, 2 or more layer networks seem to be completely
       random-like.
This is consistent with the bias-variance tradeoff that states that 
       the variance of statistical models increases with model complexity,
       while the bias is reduced.
This is also demonstrated for the two transformer networks in our bottom rows.
The images for DINOV3 appear even more random than the images for ViT-tiny.

\section{Conclusion}\label{sec:conclusion}
We have introduced two new methods for generating input images that
   produce prescribed neural network outputs.
The generated images expose the vulnerability of the underlying network
   architectures.
Currently, we are working on expanding the coverage of \textsc{BackwardPassInv}(.) 
   to use linear programming methods for generating images for neural
   networks with ReLU activations that are non-invertible.
\bibliographystyle{IEEEtran}
\bibliography{refs.bib}

\end{document}